\documentclass[10pt,twocolumn,letterpaper]{article}

\usepackage[pagenumbers]{cvpr} 

\usepackage{graphicx}
\usepackage{amsmath}
\usepackage{amssymb}
\usepackage{booktabs}
\usepackage{soul}
\usepackage{algorithm}
\usepackage{algorithmic}
\usepackage{amsthm}
\usepackage{multirow}

\makeatletter
\def\maketag@@@#1{\hbox{\m@th\normalfont\normalsize#1}}
\makeatother

\newcommand{\fbackbone}{f_\text{rep}}
\newcommand{\fproposal}{f_\text{RPN}}
\newcommand{\froi}{h_\text{RoI}}
\newcommand{\fpred}{f_\text{pred}}
\newcommand{\fSpaPool}{h_\text{SP}}
\newcommand{\proposal}{B^\text{pro}}
\newcommand{\Zbackbone}{Z^\text{rep}}
\newcommand{\Zroi}{Z^\text{RoI}}
\newcommand{\ZSpaPool}{Z^\text{SP}}

\newcommand{\numproposal}{N_\text{pro}}
\newcommand{\numEleClu}{N^{\text{c}}}

\newcommand{\numRFF}{N_\text{RFF}}
\newcommand{\Hraw}{H^\text{raw}}
\newcommand{\Wraw}{W^\text{raw}}
\newcommand{\Craw}{C^\text{raw}}
\newcommand{\Hrep}{H^\text{rep}}
\newcommand{\Wrep}{W^\text{rep}}
\newcommand{\Crep}{C^\text{rep}}
\newcommand{\Hroi}{H^\text{RoI}}
\newcommand{\Wroi}{W^\text{RoI}}
\newcommand{\Croi}{C^\text{RoI}}
\newcommand{\Nvisible}{N^\text{vis}}
\newcommand{\Ivisible}{V}

\newcommand{\calX}{\mathcal{X}}
\newcommand{\calY}{\mathcal{Y}}

\newcommand{\calB}{\mathcal{B}}
\newcommand{\calW}{\mathcal{W}}
\newcommand{\calM}{\mathcal{M}}
\newcommand{\calL}{\mathcal{L}}
\newcommand{\calE}{\mathcal{E}}
\newcommand{\bfw}{\mathbf{w}}
\newcommand{\calZbackbone}{\mathcal{Z}^\text{rep}}
\newcommand{\calZroi}{\mathcal{Z}^\text{RoI}}

\newcommand{\Ldecorr}{\calL_{\text{decorr}}}
\newcommand{\Lpred}{\calL_{\text{pred}}}
\newcommand{\Lcls}{\calL_{\text{cls}}}
\newcommand{\Lreg}{\calL_{\text{reg}}}

\newcommand\blfootnote[1]{%
  \begingroup
  \renewcommand\thefootnote{}\footnote{#1}%
  \addtocounter{footnote}{-1}%
  \endgroup
}

\usepackage[pagebackref,breaklinks,colorlinks]{hyperref}

\usepackage[capitalize]{cleveref}
\crefname{section}{Sec.}{Secs.}
\Crefname{section}{Section}{Sections}
\Crefname{table}{Table}{Tables}
\crefname{table}{Tab.}{Tabs.}

\theoremstyle{definition}
\newtheorem{problem}{Problem}

\def\cvprPaperID{2203} 
\def\confName{CVPR}
\def\confYear{2022}

\begin{document}

\title{Towards Domain Generalization in Object Detection}

\author{Xingxuan Zhang $^1$, Zekai Xu$^1$, Renzhe Xu$^1$, Jiashuo Liu$^1$, Peng Cui$^1$*, \\
Weitao Wan$^2$, Chong Sun$^2$, Chen Li$^2$ \\
$^1$Department of Computer Science, Tsinghua University \qquad $^2$ WeChat, Tencent Inc.\\
{\tt\small xingxuanzhang@hotmail.com, xuzk20@mails.tsinghua.edu.cn, xrz199721@gmail.com,} \\
{\tt\small liujiashuo77@gmail.com, cuip@tsinghua.edu.cn,  collinwan@tencent.com,} \\
{\tt\small waynecsun@tencent.com, chaselli@tencent.com}
}

\maketitle

\begin{abstract}
    Despite the striking performance achieved by modern detectors when training and test data are sampled from the same or similar distribution, the generalization ability of detectors under unknown distribution shifts remains hardly studied. 
    Recently several works discussed the detectors' adaptation ability to a specific target domain, which are not readily applicable in real-world applications since detectors may encounter various environments or situations while pre-collecting all of them before training is inconceivable. 
    In this paper, we study the critical problem, domain generalization in object detection (DGOD), where detectors are trained with source domains and evaluated on unknown target domains. 
    To thoroughly evaluate detectors under unknown distribution shifts, we formulate the DGOD problem and propose a comprehensive evaluation benchmark to fill the vacancy. Moreover, we propose a novel method named Region Aware Proposal reweighTing (RAPT) to eliminate dependence within RoI features. Extensive experiments demonstrate that current DG methods fail to address the DGOD problem and our method outperforms other state-of-the-art counterparts.

\end{abstract}

\blfootnote{*Corresponing author}


\section{Introduction}

\begin{figure}[th]
    \centering
    \includegraphics[width=\linewidth]{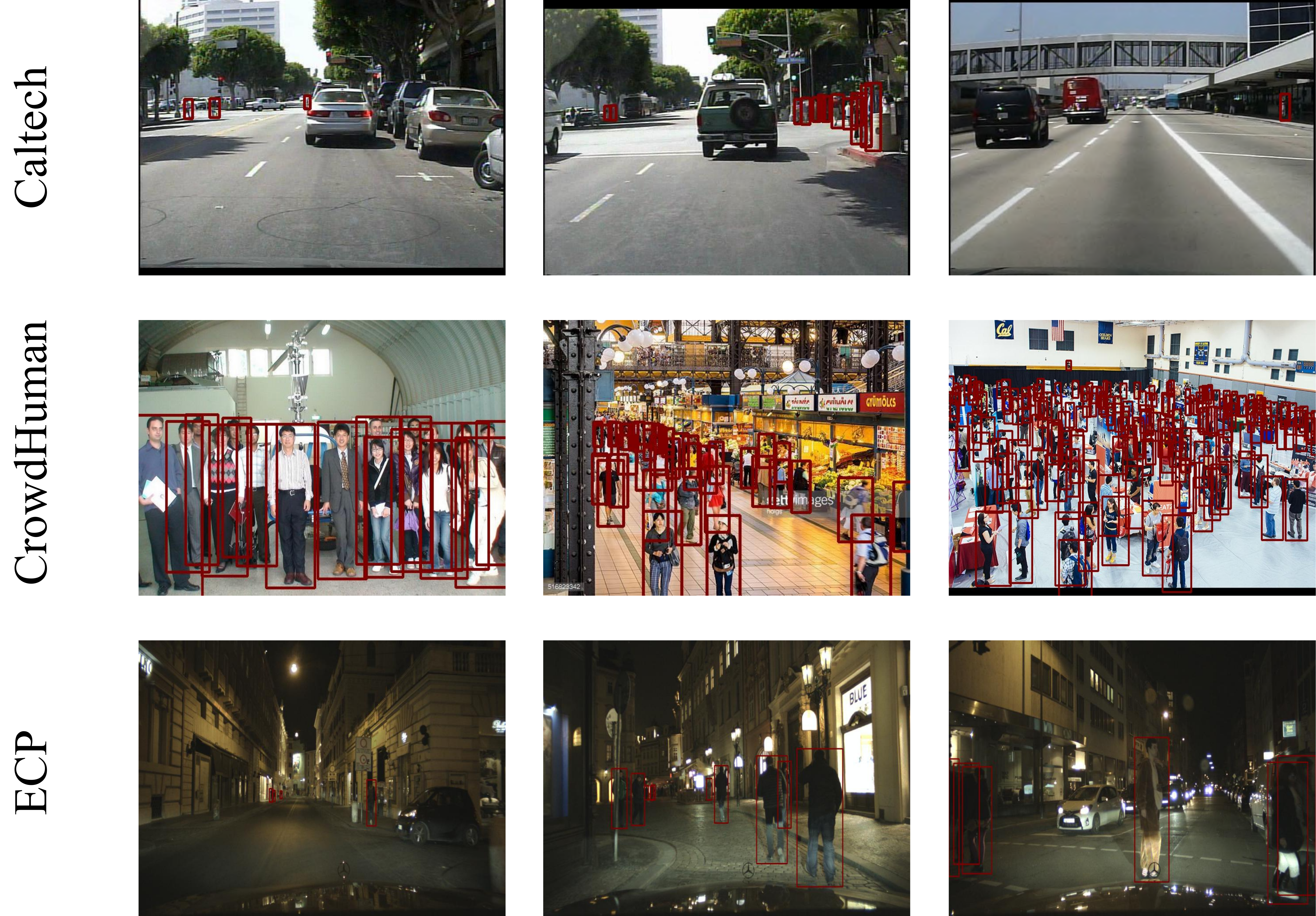}
    \caption{Comparison of Caltech \cite{2012Pedestrian}, CrowdHuman \cite{2018CrowdHuman}, and ECP \cite{Braun0EuroCity}. Clear distribution shifts are introduced by the density of people, image contexts, illumination and filming anchors.}
    \label{fig:intro}
\end{figure}

Modern detectors have shown striking performance when trained and tested with data sampled from a single benchmark \cite{zhang2020semantics}, yet many application scenarios require reliable and stable performance when encountered input data from any possible distribution including those differs from the distribution of training data \cite{shen2021towards,zaidi2022survey,liu2021survey,zou2019object}. For example, autonomous driving requires all vehicles, pedestrians and signal lights to be accurately detected under any distribution shifts caused by inconsistency in contexts, time, weather, shot angle and illumination \cite{hbaieb2019pedestrian,2021Robust}.

Since significant distribution shifts are among real-world detection data (as shown in Fig. \ref{fig:intro}) and pre-collected data can hardly involve all of the possible situations in real-world applications, the generalization ability of models to unseen distributions is critical for detectors.
Thus training and testing detectors with data sampled from a given dataset lack evaluation significance for practical applications, yet detectors may suffer from significant performance drops under simple disturbance \cite{hasan2021generalizable, 2021Robust}.
This situation brings a critical question: how to improve the generalization ability of modern detectors and how to thoroughly evaluate them towards real-world applications in unknown scenes.

Some works studied the domain adaptation (DA) problem in object detection to improve models' performance on a domain with limited category and bounding box annotations, where the distribution of test images is accessible in the training phase \cite{hsu2020progressive,Guan2021UncertaintyAwareUD,inoue2018cross}, and showed promising results. However, as mentioned above, under the circumstances where one can hardly ensure the availability of test data distribution domain adaptation methods cannot be readily applied \cite{shen2020stable}.

Despite the urgent need of the evaluation of detectors under unknown distribution shifts for real-world applications, recent literature lacks the discussions of evaluation methods and evaluation metrics for the object detection problem under distribution shifts. In this paper we formulate the problem of object detection under distribution shifts and propose a evaluation method with multiple detection datasets for it. 
To thoroughly evaluate generalization ability \cite{muandet2013domain,ganin2016domain,li2018domain}, none of the prior knowledge of test distribution is avaliable in the training phase.

Recently considerable attention has been drawn to the field of domain generalization (DG) for image recognition \cite{shen2021towards,volpi2018generalizing,shankar2018generalizing}. Specifically, in the DG literature \cite{shen2021towards,wang2021generalizing,zhou2021domain}, a category is split into multiple domains according to data source \cite{torralba2011unbiased}, context \cite{he2021towards}, or image style \cite{li2017deeper}, so that domain-related features (e.g., features that are irrelevant to categories, such as features of image style, figure resolution, etc.) vary across different domains while category relevant features remain invariant. Such a split of heterogeneous data makes it possible for a well-designed model to learn the invariant representations across domains and inhibit the negative effect from domain-related features, leading to better generalization ability under distribution shifts \cite{zhang2021deep}. 

Moreover, inspired by DG evaluation approaches \cite{li2017deeper, peng2019moment}, we propose a comprehensive cross-dataset evaluation protocol to test the generalization ability of detectors. Different from \cite{hasan2021generalizable}, which mainly focuses on adaptation between two given datasets and progressive fine-tuning, we introduce four kinds of distribution shifts via clustering several large-scale benchmarks into two groups for training and evaluation, respectively. These settings, namely \textit{classic DG}, \textit{density shift}, \textit{context shift}, and \textit{random shift}, evaluate detection robustness against diverse distribution shifts with practical meaning and significant impact. With extensive experiments, we show that RAPT improves the generalization ability of detectors under various distribution shifts. 

Essentially, model crash under distribution shifts is mainly induced by the spurious correlations between domain-related features and category labels, which are intrinsically caused by the subtle correlations between relevant and irrelevant features \cite{lake2017building, marcus2018deep, lopez2017discovering, arjovsky2019invariant, zhang2021deep}. Consider the context in street detection scenarios as an example, if pedestrians are usually on the sidewalk in the training data, there are strong spurious correlations between features of sidewalk and the label `person'. When tested with images where pedestrians are not on the sidewalk (such as on the road), detectors may yield wrong predictions. Significant drops in performance are caused by the difference in context between training and testing data \cite{zhang2021deep, he2021towards}, as shown in Section \ref{exp}.

Some recent works have proposed to decorrelate relevant features and irrelevant features with sample reweighting and shown effectiveness \cite{shen2020stable,kuang2020stable,zhang2021deep}. Nevertheless, most of them focus on simple data structures (e.g., linear models \cite{shen2020stable,kuang2020stable}) or basic tasks (e.g., image classification \cite{zhang2021deep}).

In object detection, however, one single image can yield many proposals and predictions, leading to insufficient learning of sample weights for images. 
Thus we propose a novel sample reweighting method called Region Aware Proposal reweighTing (RAPT) to learn weights for proposals to eliminate the statistical dependence between features. Since RoI features are of high dimensions, which introduces enormous calculation, RAPT clusters proposals according to their visible features and the relative position of the visible area, and learns sample weights within each cluster. In this manner, the spatial knowledge of RoI features and visible parts is leveraged for prediction while computational complexity is small enough to be ignored.

\section{Related Works}
\paragraph{Object detection and pedestrian detection}
Object detection aims to learn accurate detectors for various categories. 
There are two mainstreams for object detection: single-stage detectors \cite{liu2016ssd,redmon2016you,tian2019fcos,2017Focal,redmon2018yolov3,bochkovskiy2020yolov4} and two-stage detectors \cite{2017Faster,lin2017feature,yang2019reppoints,dai2017deformable}. The main difference of them lies in whether the proposes are filtered through second stage heads. Most of them adopt rectangular anchors for the object representation, while some of recent works proposed to present objects in terms of point sets \cite{duan2019centernet,law2018cornernet,yang2019reppoints,yang2020dense,xie2020polarmask,zhou2019bottom} and achieved comparable performance compared with rectangular anchors based ones. 
Pedestrian detection is a subfield of object detection, which is prone to be effected with distribution shifts and it requires highly reliable detectors. Thus we consider pedestrian detection as one of our evaluation scenarios.
Many modern pedestrian detection methods \cite{hosang2015taking,zhang2016far,zhang2016faster,chu2020detection,DBLP:journals/corr/abs-2002-09053,liu2018learning,DBLP:journals/corr/abs-2005-05708} are inspired by standard object detection, such as Faster RCNN \cite{2017Faster}, RetinaNet \cite{2017Focal} and Single Shot MultiBox Detector (SSD) \cite{liu2016ssd}. Recently, the occlusion problem in pedestrian detection has drawn lots of attention. To tackle this problem, penalty based methods \cite{wang2018repulsion,zhang2018occlusion} are proposed to force the detector to focus on the true target. Some methods also try to adopt visible information as external guidance to learn various occlusion patterns in crowd scenarios \cite{zhou2018bi,zhang2018occluded,pang2019mask}. 
But none of current detectors consider the robustness of pedestrian detectors under distribution shifts.

\paragraph{Domain Generalization}
With access to data from several source domains, Domain Generalization (DG) problems aim to learn models that generalize well on unseen target domains, which focuses mostly on computer vision related classification problems on the grounds that predictions are prone to be affected by disturbance on images (e.g., style, light, rotation, etc.). According to \cite{shen2021towards}, regarding to different methodological focuses, DG methods can be categorized into three branches, namely representation learning \cite{muandet2013domain,ganin2016domain,li2018domain,sun2016deep,muandet2013domain}, training strategy \cite{li2018learning,zhou2020domain,sun2019unsupervised,carlucci2019domain,zhang2021deep,zhang2021domain,xu2021stable} and data augmentation \cite{yue2019domain,peng2018sim,volpi2018generalizing,shankar2018generalizing,qiao2020learning,zhou2020learning}. Existing surveys of this field can be found in \cite{shen2021towards,wang2021generalizing,zhou2021domain}.

\paragraph{Domain Adaptation in Object Detection}
Domain adaptation (DA) aims to improve models' performance on a known target domain \cite{long2015learning,rodriguez2019domain}. Following the basic approach in DA, DA\cite{chen2018domain} and MTOR\cite{cai2019exploring} minimize domain discrepancy for object detection. SWDA \cite{saito2019strong} aligns local similar features strongly and aligns global dissimilar features weakly. IFAN \cite{zhuang2020ifan} aligns feature distributions at both image and instance levels. And other works \cite{zhao2020collaborative,xu2020exploring,Guan2021UncertaintyAwareUD,yao2021multi} also show promising performance when the distribution of the target domain is accessible in the training phase, which are not appliable in domain generalization (DG) scenarios where the information of test data is completely inaccessible. Although the domain adaptation in object detection is widely discussed, domain generalization in object detection remains unstudied.    

\section{Methods}

\begin{figure*}[ht]
    \centering
    \includegraphics[width=0.80\textwidth]{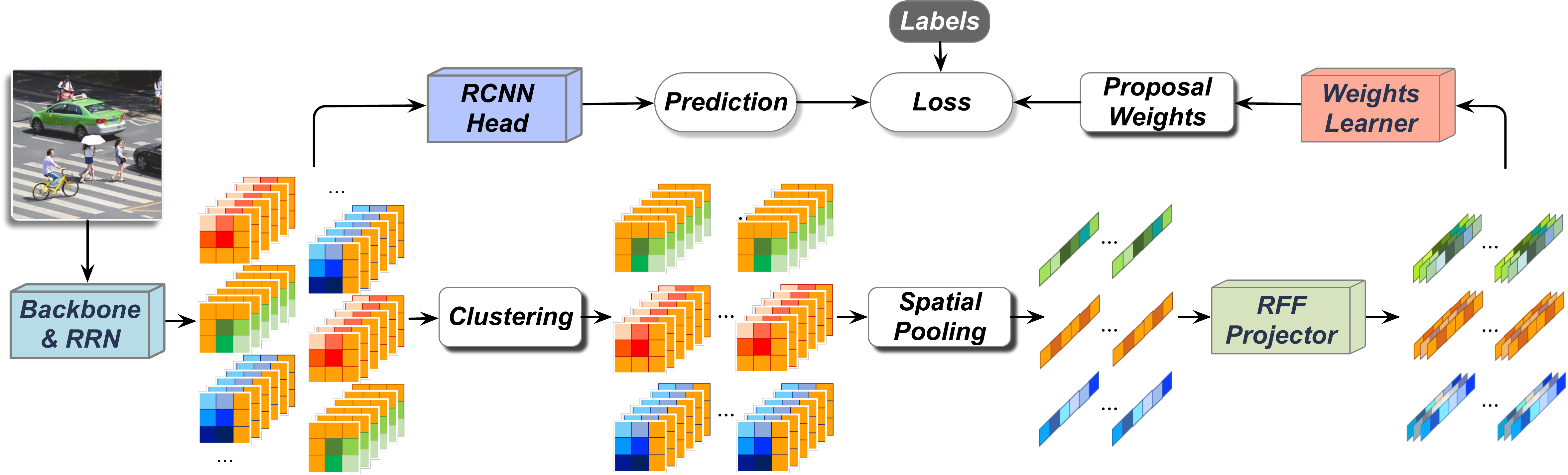}
    \vspace{-4pt}
    \caption{The overall architecture of the proposed RAPT method. The Clustering module clusters Roi features according to visible features and the relative position of visible area in the Roi. Weights Learner in the figure follows the Eqn. \ref{eq:precedure} and learns proposal weights. Loss is calculated via weighting classification loss and regression loss with proposal weights. 
    }
    \label{fig:model}
    \vspace{-8pt}
\end{figure*}

In most current anchor-based detection methods, the significance of each proposal for loss calculation is considered equally. Though several approaches \cite{shrivastava2016training, 2017Focal} highlight the contribution of hard samples to address the imbalance between categories or foreground and background, they fail to consider the possible heterogeneity and distribution shifts between training and testing data. 
We propose an effective sample reweighting method called Region-Aware Proposal reweighTing (RAPT) to improve detectors' generalization ability under distribution shifts.
RAPT eliminates statistical dependences between relevant features (i.e., features related to the target object) and irrelevant features (i.e., features that vary in different datasets), and thus spurious correlations between irrelevant features and labels are eliminated. 
As shown in Figure \ref{fig:model}, the main idea of RAPT is to consider proposals as samples and learn sample weights to reweight losses generated by corresponding proposals. We give the formulations and theoretical explanations in Section \ref{subsection:rapt} and training protocol in Section \ref{subsection:training}.

\subsection{Problem Formulation}
\paragraph{Notations}
Let $\calX \subseteq \mathbb{R}^{\Hraw \times \Wraw \times \Craw}$ denote the space of raw pixels with width $\Wraw$, height $\Hraw$, and number of channel $\Craw$, $\calY \subseteq \mathbb{R}$ denote the space of labels, 
$\calB \in \mathbb{R}^4$ denote the space of bounding boxes. A sample $Q_i = \left\{X_i, \{y_{ij}, B_{ij}\}_{j=1}^{N_i}\right\}$ with index $i$ includes image $X_i \in \calX$ and $N_i$ pairs of category labels $y_{ij} \in \calY$ and bounding boxes $B_{ij} \in \calB$.

Let $\calE$ be the set of all possible domains and a specific domain $e \in \calE$ corresponds to a distribution $P^e$ on the space of sample $Q$. Let $\calE_{\text{tr}} \subseteq \calE$ be the set of all training domains. For any training domain $e \in \calE_{\text{tr}}$, let $D_e = \left\{Q_i\right\}_{i=1}^{N_e}$ be the training data with $N_e$ samples drawn from $P^e$. In addition, there exists an unknown testing domain $e_{\text{te}} \in \calE$. 

We assume that a object detection model $f: \calX \rightarrow ((\calY \cup \{\phi\}) \times \calB)^A$ yield at most $A$ bounding boxes and category predictions from each image. For a model $f$, let $\calM(Q; f)$ be an evaluation metric on sample $Q$ (such as mAP and MR$^{-2}$) to measure the model performance.

\begin{problem} [Domain Generalization in Object Detection (DGOD)]
    Given the training data $\{D_e\}_{e \in \calE_{\text{tr}}}$ without domain labels and the evaluation metric $\calM$, the target is to maximize the evaluation metric on the unseen testing domain $e_{\text{te}}$, \textit{i.e.}, $\mathbb{E}_{Q \sim P^{e_{\text{te}}}} \left[\calM(Q;f)\right]$.
   
\end{problem}

\subsection{Overall framework}
Let $\calZbackbone \subseteq \mathbb{R}^{\Hrep \times \Wrep \times \Crep}$ denote the space of visual feature maps and $\calZroi \subseteq \mathbb{R}^{\Hroi \times \Wroi \times \Croi}$ denote the space of RoI features. Let $[K]$ denote the set $\{1, 2, \dots, K\}$.

In a anchor based detector, given an image sample $X_i$, the visual feature map $\Zbackbone_i$ is generated by a backbone representation function $\fbackbone: \calX \rightarrow \calZbackbone$. A region proposal network (RPN) $\fproposal: \calZbackbone \rightarrow \Delta_\calB$ is then adopted to generate a distribution on $\calB$ and samples $\numproposal$ proposals $\{\proposal_{ij} \in \calB\}_{j=1}^{\numproposal}$ from the distribution. Here $\Delta_\calB$ denotes the space of all distributions on $\calB$. 

To regularize the size of features input into the following heads in modern detectors, a RoI pooling function $\froi: \calB \times \calZbackbone \rightarrow \calZroi$ , such as RoI pooling \cite{girshick2015fast}, aligned RoI pooling \cite{he2017mask} and deformable RoI pooing \cite{dai2017deformable} is adopted to generate the RoI features $\Zroi_{ij}$ of each proposal for the final predicting function $\fpred: \calZroi \rightarrow \calY \times \calB$ to generate bounding box regression $\hat{y}_{ij}$ and $\hat{B}_{ij}$ of proposals.

\subsection{Region-Aware Proposal Weights Learning}
\label{subsection:rapt}
Though sample reweighting based decorrelation methods show effectiveness in regression and classification tasks \cite{shen2020stable,zhang2021deep}, applying proposal reweighting in detectors suffers major problems.

RoI features $\Zroi$ still maintain the spatial dimension, where similar features may lead to variant instructions for class and bounding box prediction (e.g., human head related features give different instructions to the regression of the bounding box when detected on the right or left of the feature map). Thus eliminating dependences between category relative and irrelevant features requires decorrelations between features within each bin of the RoI feature.

Since decorrelating all the features inside each bin of RoI features of each input image introduces excessive calculation, we propose a region-based proposal reweighting method to effectively learn weights for proposals. 
Given similar features from the same visible position of different proposals should give similar instructions for the classification and regression, we consider proposals with similar visible area locations as a cluster and learn weights for all of the samples within the cluster to decorrelate their RoI features. We present our method in detail as follows. 

\subsubsection{Clustering}
We first cluster the proposals by their bounding boxes, RoI features, and visible information with a clustering function $g: \calB \times \calZroi \times \{0,1\}^{\Hroi \times \Wroi}\rightarrow [K]$.
The cluster of a proposal is given by $G_{ij} \triangleq g\left(\proposal_{ij}, \Zroi_{ij}, \Ivisible_{ij}\right)$.
Here $\Ivisible_{ij} \in \{0, 1\}^{\Hroi \times \Wroi}$ denotes the visible regions of the proposal $\proposal_{ij}$ and $\left(\Ivisible_{ij}\right)_{hw} \in \{0, 1\}$ indicates whether the region in $\proposal_{ij}$ with coordinates $(h, w)$ is visible. For datasets without the annotation of visible areas, $\Ivisible_{ij} \equiv 1^{\Hroi \times \Wroi}$.

We use k-means clustering \cite{macqueen1967some} to learn the clustering function $g$. Let $\numEleClu_k$ denotes the number of proposals in cluster $k$, \textit{i.e.}, $\numEleClu_k = \sum_{i=1}^N\sum_{j=1}^{\numproposal} \mathbb{I}\left[G_{ij}=k\right]$.

\subsubsection{RFF-based decorrelation}
We propose to decorrelate RoI features after an extra spatial pooling with random fourier features (RFF) \cite{rahimi2007random}.

\paragraph{Spatial pooling}
In each group, we simply adopt spatial pooling to reduce the feature dimension given that no spatial difference remains after clustering. Specifically, we squeeze spatial dimensions of given RoI features in visible areas with spatial pooling function $\fSpaPool: \calZroi \rightarrow \mathbb{R}^{\Croi}$, which can be written as
\begin{equation}
    \small
    \ZSpaPool_{ij} = \fSpaPool\left(\Zroi_{ij}\right) = \frac{1}{\Nvisible_{ij}}\sum_{h = 1}^{\Hroi}\sum_{w=1}^{\Wroi} \left(\Ivisible_{ij}\right)_{hw}\left(\Zroi_{ij}\right)_{hw}.
\end{equation}
Here $\left(\Zroi_{ij}\right)_{hw}$ is a $\Croi$-dimensional vector which represents a bin in $\Zroi_{ij}$ with coordinates $(h, w)$.
$\left(\Nvisible_{ij}\right)_{hw}$ represents the number of visible regions in $\Zroi_{ij}$, \textit{i.e.}, $\Nvisible_{ij} = \sum_{h=1}^{\Hroi}\sum_{w=1}^{\Wroi}(\Ivisible_{ij})_{hw}$.

\paragraph{Sample reweighting}
To eliminate the spurious correlation between domain-related features and category discriminating features, we seek to decorrelate all the features within $\ZSpaPool$. There are many approaches for feature decorrelation, such as Hilbert-Schmidt Independence Criterion (HSIC) \cite{gretton2008kernel} and independent component analysis (ICA) \cite{gretton2005measuring}, yet the calculation of them requires noticeable computational cost which grows as the batch size of training data increases, so it is inapplicable to training deep models on large datasets. Thus we adopt Random Fourier Features (RFF) \cite{rahimi2008random} to capture the non-linear relationships between variables for the feature statistically independence. 

Specifically, we learn sample weights to decorrelate RoI features inside each cluster by minimizing the following loss function.

\begin{equation}\label{eq:learning_weight}
    \small
    \Ldecorr\left(\bfw; \fbackbone, \fproposal, g\right) \triangleq \sum_{k=1}^K\sum_{1 \le p < q \le \Croi} \left\Vert\hat{\Sigma}_{pq|k;\bfw}\right\Vert_F^2.
\end{equation}
The space of all possible sample weights $\calW$ is given by
\begin{equation}\label{eq:weighting-class}
    \footnotesize
    \calW = \left\{\bfw \in \mathbb{R}^{N \times \numproposal}_+ \Bigg| \forall k \in [K], \sum_{i=1}^N\sum_{j=1}^{\numproposal} w_{ij}\mathbb{I}[G_{ij}=k] = \numEleClu_k \right\},
\end{equation}
which is a normalization operation that constraints the sum of proposal weights for each cluster to the number of samples within. In addition, $\hat{\Sigma}_{pq|k;\bfw}$ is the partial cross-covariance matrix between the $p$-th and $q$-th dimension of the RoI feature after extra spatial pooling $\fSpaPool$, \textit{i.e.}, $\ZSpaPool_{ijp}$ and $\ZSpaPool_{ijq}$, in cluster $k$ under sample weights $\bfw$, which is given by
\begin{equation}\label{eq:cross-covariance}
    \small
    \begin{aligned}
        & \hat{\Sigma}_{pq|k;\bfw} = \frac{1}{\numEleClu_k - 1} \sum_{i=1}^N \sum_{j=1}^{\numproposal} \mathbb{I}\left[G_{ij}=k\right] \cdot \\
        & \,\,\left(w_{ij}\mathbf{r}\left(\ZSpaPool_{ijp}\right) - \bar{\mathbf{r}}_{\bfw,k}\right)^T \left(w_{ij}\mathbf{s}\left(\ZSpaPool_{ijq}\right) - \bar{\mathbf{s}}_{\bfw,k}\right).
    \end{aligned}
\end{equation}
Here
\begin{equation}\label{eq:rff_transform}
    \small
    \left\{
    \begin{aligned}
        \mathbf{r}(\cdot) & = \left(r_1(\cdot), r_2(\cdot), \dots, r_{\numRFF}(\cdot)\right), r_l(\cdot) \in \mathcal{H}_\text{RFF}, \forall l,\\
        \mathbf{s}(\cdot) & = \left(s_1(\cdot), s_2(\cdot), \dots, s_{\numRFF}(\cdot)\right), s_l(\cdot) \in \mathcal{H}_\text{RFF}, \forall l,
    \end{aligned}
    \right.
\end{equation}
are Random Fourier Features from the following function space
\begin{equation}
    \small
    \begin{aligned}
        \mathcal{H}_\text{RFF} = \big\{ & h: x \rightarrow \sqrt{2}\cos(\omega x + \phi) \mid \\
        & \omega \sim N(0, 1), \phi \sim \text{Uniform}(0, 2 \pi)\big\}.
    \end{aligned}
\end{equation}
$\bar{\mathbf{r}}_{w,k}$ and $\bar{\mathbf{s}}_{w,k}$ are weighted means of the corresponding functions, \textit{i.e.},
\begin{equation}
    \small
    \left\{
    \begin{aligned}
        \bar{\mathbf{r}}_{\bfw,k} & = \frac{1}{\numEleClu_k}\sum_{i=1}^N\sum_{j=1}^{\numproposal} w_{ij}\mathbb{I}\left[G_{ij}=k\right]\mathbf{r}\left(\ZSpaPool_{ijp}\right), \\
        \bar{\mathbf{s}}_{\bfw,k} & = \frac{1}{\numEleClu_k}\sum_{i=1}^N\sum_{j=1}^{\numproposal} w_{ij}\mathbb{I}\left[G_{ij}=k\right]\mathbf{s}\left(\ZSpaPool_{ijq}\right). \\
    \end{aligned}
    \right.
\end{equation}

As a result, $\Ldecorr$ in Equation \ref{eq:learning_weight} can effectively decorrelate features in each cluster.

\subsection{Training procedure}

\renewcommand{\algorithmicrequire}{\textbf{Input:}}
\renewcommand{\algorithmicensure}{\textbf{Output:}}
\begin{algorithm}[t]
    \caption{\emph{Training procedure of \setlength{\parindent}{4em}RAPT}}
    \label{alg}
    \begin{algorithmic}[1]\label{alg:RAPT}
        \FOR{epoch $\leftarrow$ 1 to \#EPOCH}
            \FOR{batch $\leftarrow$ 1 to \#BATCH}
                \STATE Use K-means to learn the clustering function $g$ in the batch
                \FOR{epoch\_decorr  $\leftarrow$ 1 to \#EPOCH\_DECORR}
                    \STATE Optimize sample weights $\bfw$ in the current batch as shown in Equation \ref{eq:learning_weight}
                \ENDFOR
                \STATE Learn predicting functions $\fbackbone$, $\fproposal$, $\fpred$ under learned weights $\bfw$ as shown in Equation \ref{eq:prediction-loss}
            \ENDFOR
        \ENDFOR
        \ENSURE predicting functions $\fbackbone$, $\fproposal$, $\fpred$
    \end{algorithmic}
\end{algorithm}

\label{subsection:training}

\begin{table*}[ht]
    \centering
    \caption{Results (mAP) of detectors on the classic DG setting for general object detection. All detectors are reimplemented with ResNet-50 \cite{he2016deep} pretrained on Imagenet \cite{deng2009imagenet} as the backbone. The title of each column indicates the dataset models are tested on while the other datasets are used as training data. The best results of all methods are highlighted with the bold font.}
    \scalebox{0.85}{
    \begin{tabular}{c|cccc|c}
        \toprule
        Method & BDD100k & Cityscapes & Sim10k & KITTI & Mean  \\
        \midrule
        Faster RCNN  & 0.371 & 0.452 & 0.303 & 0.484 & 0.403 \\
        RetinaNet   & \textbf{0.414} & 0.439 & 0.279 & 0.454 & 0.397  \\
        Jigen + Faster RCNN   & 0.374 & 0.451 & 0.295 & 0.485 & 0.401 \\
        RSC + Faster RCNN    & 0.356 & 0.422 & 0.297 & 0.472 & 0.387 \\
        StableNet + Faster RCNN  & 0.373 & 0.440 & 0.297 & 0.486 & 0.399 \\
        RAPT(ours) + Faster RCNN  & 0.383 & \textbf{0.459} & \textbf{0.322} & \textbf{0.490} & \textbf{0.414} \\
        \bottomrule
    \end{tabular}}
    \label{tab:general}
\end{table*}

\begin{table*}[ht]
    \centering
    \caption{Results of detectors trained on CrowdHuman and ECP, and tested on CityPersons, Caltech and WiderPedestrian. The title of each column indicates the tested subset. For details about the number of runs and fonts, see Table \ref{tab:general}.}
    \resizebox{\textwidth}{!}{
    \begin{tabular}{c|c|cccc|cccc|c}
        \toprule
         Method & Training Data & \multicolumn{4}{c|}{CityPersons} & \multicolumn{4}{c|}{Caltech}  & \multicolumn{1}{c}{WiderPedestrian} \\
         \cmidrule{3-11} 
         & & Reasonable & Small & Heavy & All & Reasonable & Small & Heavy & All & All  \\
        \midrule
    
        CrowdDet \cite{chu2020detection} & CrowdHuman+ECP & 0.2336 & 0.3859 & 0.5170 & 0.4841 & 0.2633 & 0.3212	& 0.6722 & 0.6621 & 0.7054 \\
        IterDet  \cite{DBLP:journals/corr/abs-2005-05708} & CrowdHuman+ECP & 0.4766 & 0.6565	& 0.7920 & 0.7326 & 0.4456 & 0.4396	& 0.8381 & 0.7710 & 0.8112  \\
        ALFNet \cite{liu2018learning} & CrowdHuman+ECP & 0.2446 & 0.3791 & 0.5408 & 0.5166 & 0.2422 & 0.3435 & 0.7067 & 0.6829 & 0.7379 \\
        ACSP \cite{DBLP:journals/corr/abs-2002-09053} & CrowdHuman+ECP & \textbf{0.2118} & 0.4061	& 0.5685 & 0.4872 & 0.2809 & 0.3466	& 0.6369 & 0.6765 & 0.7652 \\
        Det-AdvProp \cite{2021Robust} &  CrowdHuman+ECP & 0.2576 & 0.4051 & 0.5426 & 0.5391 & 0.2799 & 0.3537 & 0.7470 & 0.6851 & 0.7426 \\
        LLA \cite{2021LLA} & CrowdHuman+ECP & 0.2610	& \textbf{0.2640} & 0.5445 & 0.4956 & 0.2394	& 0.2904	& 0.6621 & 0.6361 & 0.6765 \\
        RetinaNet \cite{2017Focal} & CrowdHuman+ECP & 0.2621 & 0.4483 & 0.5082 & 0.5096 & 0.2422	& 0.2831 & 0.6578 & 0.6492 & 0.6940 \\
        Cascade RCNN \cite{0Cascade} & CrowdHuman+ECP & 0.3455 & 0.5051 & 0.6970 & 0.5696 & 0.3400	& 0.3965 & 0.7721 & 0.7164 & 0.7604 \\
        \midrule
        Faster RCNN \cite{2017Faster} & CrowdHuman+ECP & 0.2211 & 0.3309 & 0.4971 & 0.4705 & 0.2165 & 0.2612	& 0.6376 & 0.6162 & 0.6852  \\
        RAPT (ours) + FRCNN & CrowdHuman+ECP  & 0.2170 & 0.3438 & \textbf{0.4724} & \textbf{0.4583} & \textbf{0.1900}  & \textbf{0.2554} & \textbf{0.6280} & \textbf{0.6126} & \textbf{0.6533} \\
        \bottomrule
    \end{tabular}}
    \label{tab:dense1}
\end{table*}

The prediction loss for the representation function $\fbackbone$, region proposal network $\fproposal$, and predicting function $\fpred$ under weights $w$ is given by
\begin{equation} \label{eq:prediction-loss}
    \small
    \begin{aligned}
        & \Lpred(\fbackbone, \fproposal, \fpred; w) \\
        \triangleq & \sum_{i=1}^N\sum_{j=1}^{\numproposal}w_{ij}\left(\Lcls(\bar{y}_{ij}, \hat{y}_{ij}) + \Lreg(\bar{B}_{ij}, \hat{B}_{ij})\right).
    \end{aligned}
\end{equation}
Here $\bar{y}_{ij}$ and $\bar{B}_{ij}$ are the corresponding ground truth label and bounding box \textit{w.r.t.} the output $(\hat{y}_{ij}, \hat{B}_{ij})$. $\Lcls(\cdot, \cdot)$ is a standard classification loss and we adopt the binary cross entropy loss in this RAPT. $\Lreg(\cdot, \cdot)$ measures the error between the ground truth bounding box $\bar{B}_{ij}$ and the predicted one $\hat{B}_{ij}$. We adopt smooth L1 loss in practice.

Our algorithm iteratively optimize detection functions (including representation function $\fbackbone$, region proposal network $\fproposal$, and predicting function $\fpred$), clustering function $g$, and sample weights $w$ as follows:
\begin{equation}
\label{eq:precedure}
    \footnotesize
    \left\{
    \begin{aligned}
        & \fbackbone^{(t+1)}, \fproposal^{(t+1)}, \fpred^{(t+1)} = \underset{\fbackbone, \fproposal, \fpred}{\arg \min} \Lpred\left(\fbackbone, \fproposal, \fpred; w^{(t)}\right), \\
        & g^{(t+1)} = \text{K-means}(K, \fbackbone, \fproposal), \\
        & w^{(t+1)} = \underset{w \in \calW}{\arg \min}\, \Ldecorr\left(w; \fbackbone^{(t+1)}, \fproposal^{(t+1)}, g^{(t+1)}\right).
    \end{aligned}
    \right.
\end{equation}
Here $f_{\cdot}^{(t)}$, $g^{(t)}$, $w^{(t)}$ means the functions and sample weights at time stamp $t \in [T]$. $\text{K-means}(K, \fbackbone, \fproposal)$ denotes the output function of the K-means algorithm that categorizes the features $\left(\proposal_{ij}, \Zroi_{ij}, \Ivisible_{ij}\right)$ generated by $\fbackbone$ and $\fproposal$ into $K$ clusters. Initially, $w^{(0)} = (1, 1, \dots, 1)$.

Equations listed above require to learn a weight for any possible proposals in the training data, yet in practice, only part of the proposals are observed in each batch with SGD \cite{bottou201113} as the optimizer. As a result, we slightly change the equations to calculate loss functions in each batch. A detailed training procedure of our proposed method RAPT is shown in Algorithm \ref{alg:RAPT}.

\section{Experiments}
\label{exp}
Traditional detectors are usually evaluated with i.i.d. data, i.e., the training and test data are from a single dataset sharing the same distribution. 
Most current detection datasets do not consider distribution shifts or only cluster data into two domains which are insufficient for DG evaluation. The combination of different datasets for training and test can introduce significant distribution shifts since the density of objects, image contexts, illumination and filming anchors across different datasets vary largely, especially in pedestrian detection datasets.
To evaluate the robustness of pedestrian detectors under distribution shifts and explore the effectiveness of cross-dataset data augments, such as whether open-world detection datasets help detection in autonomous driving scenarios, we propose four novel evaluation settings to benchmark detectors under distribution shifts between training and test data in both general detection and pedestrian detection scenarios. 

\subsection{Benchmark for Object Detection in Domain Generalization}

\paragraph{Datasets.} To thoroughly evaluate current detectors under distribution shifts, we adopt 5 large-scale general detection datasets and 6 pedestrian detection datasets for extensive settings.

\begin{table}[t]
    \centering
    \caption{Instance density of each dataset}
    \resizebox{\linewidth}{!}{
    \begin{tabular}{c|c|c}
        \toprule
        \textbf{Dataset} & \textbf{\# of objects/img} & \textbf{\# of overlaps/img} \\
        \midrule
        COCO \cite{DBLP:journals/corr/LinMBHPRDZ14} & 9.34 & 0.015 \\
        Caltech \cite{2012Pedestrian} & 4.92 & 0.08 \\
        Citypersons \cite{zhang2017citypersons} & 6.47 & 0.32 \\
        EuroCity Persons (ECP) \cite{Braun0EuroCity} & 11.65 & 0.63 \\
        CrowdHuman \cite{2018CrowdHuman} & 22.64 & 2.40 \\
        WiderPedestrian \cite{DBLP:journals/corr/abs-1902-06854} & 6.05 & 0.09 \\
        \bottomrule
    \end{tabular}}
    \label{tab:instance-density}
\end{table}

\begin{table}[t]
    \centering
    \caption{Evaluation splits.}
    \resizebox{0.9\linewidth}{!}{
    \begin{tabular}{c|cccc}
        \toprule
        \textbf{Setting} & Resonable & Small & Heavy & All \\
        \midrule
        \textbf{Height} & $[50, \infty)$ & $[50, 75]$ & $[50, \infty)$ & $[20, \infty)$ \\
        \midrule
        \textbf{Visibility} & $[0.65, 1]$ & $[0.65, 1]$ & $[0.2, 0.65]$ & $[0.2, 1]$ \\
        \bottomrule
    \end{tabular}}
    \label{tab:occlusion}
\end{table}

\begin{table*}[ht]
    \centering
    \caption{Results of detectors trained on sparse datasets and tested on crowd ones. For details about the number of runs, meaning of column titles and fonts, see Table \ref{tab:dense1}.}
    \scalebox{0.75}{
    \begin{tabular}{c|c|cccc|cccc}
        \toprule
        \multirow{2}{*}{Model} & \multirow{2}{*}{Training Data} & \multicolumn{4}{c|}{CrowdHuman} & \multicolumn{4}{c}{ECP}   \\
        \cmidrule{3-10} 
         & & Reasonable & Small & Heavy & All & Reasonable & Small & Heavy & All   \\
        \midrule
        IterDet   & City+Caltech+Wider & 0.6135 & 0.4728 & 0.9126 & 0.8172 & 0.3724	& 0.5431 & 0.8101 & 0.6384   \\
        ALFNet  & City+Caltech+Wider & 0.5252 & 0.4473 & 0.8912 & 0.7917 & 0.2453 & 0.3846 & 0.7077 &  0.5755  \\
        ACSP  & City+Caltech+Wider & 0.5876 & 0.4278 & 0.8842 & 0.7981 & 0.2518	& 0.3765 & 0.7483 & 0.5782  \\
        Det-AdvProp  & City+Caltech+Wider & 0.5236 & 0.4597 & 0.8911 & 0.7573 & 0.2305 & 0.3293 & 0.6719 & 0.5630  \\
        LLA  & City+Caltech+Wider & 0.5280 & 0.4315 & 0.8900 & 0.7793	& 0.2342 & 0.3533 & 0.7017	& 0.5340   \\
        RetinaNet  & City+Caltech+Wider & 0.5884 & 0.4683 & 0.8770 & 0.8175	& 0.2619 & 0.3978 & 0.6878 & 0.5680  \\
        Cascade RCNN & City+Caltech+Wider & 0.6631 & 0.5164	& 0.9108 & 0.8190 & 0.3781 & 0.4610	& 0.8561	& 0.6370 \\
        \midrule
        Faster RCNN  & City+Caltech+Wider & 0.5080 & 0.4195 & 0.8635 & 0.7600 & 0.2493	& 0.3703 & 0.7182 & 0.5549   \\
        RAPT (ours) + FRCNN  & City+Caltech+Wider  & 0.4970 & \textbf{0.3938} & 0.8748 & 0.7480 & \textbf{0.1987}  & 0.2796 & 0.6673 & 0.5238 \\
        \midrule
        CrowdDet  & City+Caltech+Wider & 0.4907 & 0.4058 & \textbf{0.8642} & 0.7443 & 0.2106	& 0.3008 & 0.6778 & 0.5031  \\
        RAPT (ours) + CrowdDet  & City+Caltech+Wider  & \textbf{0.4715} & 0.3980 & 0.8698 & \textbf{0.7305} & 0.1998  & \textbf{0.2650} & \textbf{0.6604} & \textbf{0.4927}  \\
        \bottomrule
    \end{tabular}}
    \label{tab:dense2}
\end{table*}

\begin{table*}[ht]
    \centering
    \caption{Results of detectors trained on open-world datasets and tested on autonomous driving ones. For details about the number of runs, meaning of column titles and fonts, see Table \ref{tab:dense1}.}
    \resizebox{\textwidth}{!}{
    \begin{tabular}{c|c|cccc|cccc|cccc}
        \toprule
         \multirow{2}{*}{Model} & \multirow{2}{*}{Training} & \multicolumn{4}{c|}{ECP} & \multicolumn{4}{c|}{Caltech}  & \multicolumn{4}{c}{CityPersons} \\
         \cmidrule{3-14} 
         & & Reasonable & Small & Heavy & All & Reasonable & Small & Heavy & All & Reasonable & Small & Heavy & All  \\
        \midrule
        CrowdDet & C+W & 0.2293	& 0.3802 & 0.7276 & 0.5364 & 0.2292	& 0.2786 & 0.6293 & 0.6299 & 0.2984 & 0.4940 & 0.6493	& 0.5449\\
        IterDet & C+W & 0.4057 & 0.6111 & 0.8268 & 0.6614 & 0.2727	& 0.3271 & 0.7054 & 0.6895 & 0.4051 & 0.5984 & 0.7650 & 0.6272\\
        ALFNet & C+W & 0.2569 & 0.4146 & 0.7285 & 0.5636 & 0.2552  & 0.2896 & 0.6418 & 0.6489 & 0.3036 & 0.4662 & 0.6317  & 0.5559 \\
        ACSP & C+W & 0.2780	& 0.4219 & 0.7644 & 0.5808 & 0.2225	& 0.2581 & 0.6451 & 0.6404 & 0.2967 & 0.5584 & 0.6962	& 0.5669\\
        LLA & C+W & 0.2352 & 0.3761	& 0.7451 & 0.5428 & \textbf{0.2155} & 0.2628 & 0.6357 & 0.6235	& 0.3190 & 0.4557	& 0.6733 & 0.5632\\
        RetinaNet & C+W & 0.2576 & 0.4079 & 0.7072 & 0.5659	& 0.2156 & \textbf{0.2512} & 0.6237 & 0.6312	& 0.3169 & 0.5114 & 0.6250	& 0.5654 \\
        Cascade RCNN & C+W & 0.3390 & 0.4632 & 0.8298 & 0.6140	& 0.2734 & 0.3129 & 0.6849 & 0.6716	& 0.3608 & 0.5243	& 0.7518 & 0.5933\\
        \midrule
        Faster RCNN & C+W & 0.2162	& 0.3499 & \textbf{0.7035} & 0.5239 & 0.2337	& 0.2826 & 0.6206 & 0.6297 & \textbf{0.2912}	& 0.4679 & 0.6139 & 0.5364\\
        RAPT (ours) + FRCNN & C+W  & \textbf{0.1982} & \textbf{0.3209} & 0.7144 & \textbf{0.4993} & 0.2282  & 0.2635 & \textbf{0.6122} & \textbf{0.6134} & 0.2942 & \textbf{0.4414} & \textbf{0.5849} & \textbf{0.5193} \\
        \bottomrule
    \end{tabular}}
    \label{tab:context1}
\end{table*}

\begin{table*}[ht]
    \centering
    \caption{Results of detectors trained on autonomous driving datasets and tested on open-world ones. For details about the number of runs, meaning of column titles and fonts, see Table \ref{tab:dense1}.}
    \scalebox{0.86}{
    \begin{tabular}{c|c|cccc|c}
        \toprule
        \multirow{2}{*}{Model} & \multirow{2}{*}{Training Data} & \multicolumn{4}{c|}{CrowdHuman} & \multicolumn{1}{c}{WiderPedestrian} \\
        \cmidrule{3-7} 
        & & Reasonable & Small & Heavy & All & All  \\
        \midrule
        CrowdDet & ECP+Caltech+City & 0.6757 & 0.6178 & 0.9251 & 0.8329	& 0.8084\\
        IterDet & ECP+Caltech+City & 0.9217 & 0.7482 & 0.9683 & 0.9541	& 0.9196 \\
        ALFNet & ECP+Caltech+City & 0.8453 & 0.6719 & 0.9295  & 0.9012 & 0.8185 \\
        ACSP & ECP+Caltech+City & 0.8934 & 0.6645 & 0.9633 & 0.9472 & 0.8063\\
        LLA  & ECP+Caltech+City & 0.6527 & 0.5800	& 0.9141 & 0.8232 & 0.8191 \\
        RetinaNet  & ECP+Caltech+City & 0.6868 & 0.6569 & 0.9247	& 0.8523 & 0.8030\\
        Cascade RCNN  & ECP+Caltech+City & 0.8347 & 0.7566 & 0.9681	& 0.9113 & 0.8823 \\
        \midrule
        Faster RCNN  & ECP+Caltech+City & \textbf{0.6451} & 0.5776 & 0.9122 & 0.8164	& 0.8003 \\
        RAPT (ours) & ECP+Caltech+City  & 0.6476   & \textbf{0.5619} & \textbf{0.9013} & \textbf{0.8078} & \textbf{0.7852} \\
        \bottomrule
    \end{tabular}}
    \label{tab:context2}
\end{table*}

\begin{table*}[ht]
    \centering
    \caption{Results of detectors evaluated under random distribution shifts. For details about the number of runs, meaning of column titles and fonts, see Table \ref{tab:dense1}.}
    \resizebox{\textwidth}{!}{
    \begin{tabular}{c|c|cccc|cccc|c|c}
        \toprule
        \multirow{2}{*}{Model} & \multirow{2}{*}{Training Data} & \multicolumn{4}{c|}{CrowdHuman} & \multicolumn{4}{c|}{Caltech} & \multicolumn{1}{c|}{WiderPedestrian} & \multicolumn{1}{c}{COCO} \\
        \cmidrule{3-12} 
        & & Reasonable & Small & Heavy & All & Reasonable & Small & Heavyy & All & All & All  \\
        \midrule
        CrowdDet & ECP+City & 0.5813 & 0.4494 & 0.8721 & 0.7677 & 0.3452 & 0.4114 & 0.7293 & 0.7112 & 0.7375 & 0.7869\\
        IterDet & ECP+City & 0.8884 & 0.7277 & 0.9477 & 0.9463	& 0.3852 & 0.4554 & 0.7872 & 0.7742	& 0.8876 & 0.9774 \\
        ALFNet & ECP+City & 0.6005 & 0.5129 & 0.8715 & 0.8063 & 0.3595 & 0.4278 & 0.8219 & 0.7452 & 0.8251 & 0.8051 \\
        LLA & ECP+City & 0.5884 & 0.4660 & 0.8869	& 0.7841 & \textbf{0.2849} & 0.3610 & 0.7304 & \textbf{0.6855} & 0.7516 & 0.7707	 \\
        RetinaNet & ECP+City & 0.5665 & 0.4521 & 0.8567	& 0.7866 & 0.3281 & 0.4027 & 0.7428	& 0.7135 & 0.7377 & \textbf{0.7521} \\
        Faster RCNN & ECP+City & 0.5668 & 0.4544 & \textbf{0.8513} & 0.7643	& 0.3330 & 0.4082 & 0.7196 & 0.7109 & 0.7332 & 0.7844 \\
        RAPT (ours)+FRCNN & ECP+City  & \textbf{0.5639}   & \textbf{0.4100} & 0.8606 & \textbf{0.7526} & 0.3051  & \textbf{0.3587} & \textbf{0.6952} & 0.6937 & \textbf{0.7251} & 0.7742 \\
        \bottomrule
    \end{tabular}}
    \label{tab:comp_random}
\end{table*}

\textbf{COCO} \cite{DBLP:journals/corr/LinMBHPRDZ14} is a large-scale object detection, segmentation, and captioning dataset that contains annotations of 80 different classes. In this task. We only focus on the testing results on people (label equals 1) of the validation set.

\textbf{BDD100k} \cite{Xu2017EndtoEndLO} is a large-scale, diverse dataset for autonomous driving and consists of 100,000 videos. It covers different weather conditions, including sunny, overcast, and rainy, as well as different times of day including daytime and nighttime.

\textbf{Cityscapes} \cite{Cordts2016Cityscapes} consists of 2975 training samples and 500 validation samples for semantic understanding of urban street scenes. The images are taken from 50 cities on daytime with totally 30 object categories. 

\textbf{Sim10k} \cite{johnson2016driving} is a synthetic dataset containing 10,000 images. It is generated based on the video game Grand Theft Auto V (GTA V) by incorporating photo-realistic computer images from a simulation engine to rapidly generate annotated data that can be used for the training of machine learning algorithms.

\textbf{KITTI} \cite{Geiger2012CVPR} for object detection and object orientation estimation benchmark consists of 7481 training images and 7518 test images, comprising a total of 80,256 labeled objects. All images were taken on the street by cameras on cars. 

\textbf{Caltech} \cite{2012Pedestrian} consists of approximately 10 hours of 640x480 30Hz video taken from a vehicle driving through regular traffic in an urban environment. About 250,000 frames (in 137 approximately minute long segments) with a total of 350,000 bounding boxes and 2300 unique pedestrians were annotated.
All experiments on Caltech are conducted under the new annotations provided by \cite{zhang2016far}.

\textbf{Citypersons} \cite{zhang2017citypersons} is a subset of Cityscapes\cite{Cordts2016Cityscapes} which only consists of person annotations and exhibits more diversity when compared with Caltech\cite{2012Pedestrian}. There are 2975 images for training, 500 and 1575 images for validation and testing.

\textbf{EuroCity Persons (ECP)} \cite{Braun0EuroCity} is a new dataset that is recorded in 31 different cities across 12 countries in Europe. It has 40,217 images for daytime and 7118 images for nighttime (thus referred to as ECP daytime and ECP night-time). Total annotated bounding boxes are over 200K. As mentioned in ECP, for the sake of comparison with other approaches, all experiments and comparisons are made on the daytime ECP. ECP surpasses Caltech and Citypersons a lot in terms of diversity and difficulty.

\textbf{CrowdHuman} \cite{2018CrowdHuman} is a benchmark dataset to better evaluate detectors in crowd scenarios which contain 15,000, 4,370, and 5,000 images for training, validation, and test, respectively. We test the results on the validation dataset under the same settings for a fair comparison.

\textbf{WiderPedestrian} \cite{DBLP:journals/corr/abs-1902-06854} is a non-traffic related recent benchmark dataset that contains 43,378 and 5,000 images for training and validation, we only compare the mean miss rate on the validation set due to the lack of visible bounding box annotations.

\paragraph{Benchmark.} 
For general object detection, we consider the classic evaluation method in DG, leave-one-out evaluation \cite{li2017deeper}. Specifically, we consider 4 datasets, namely BDD100K, Cityscapes, Sim10K and KITTI as 4 domains, and train detectors on three of them while test on the last for each run. 
For pedestrian detection scenarios, we consider three kinds of simple yet common distribution shifts in real pedestrian detection applications and split these datasets into subgroups, resulting in three corresponding evaluation settings, namely density shift, context shift, and random shift. 
For each setting, we divide datasets into training split and testing split following the corresponding rule.
For each dataset in the training split, we train detectors with its training subset, while for each dataset in the testing split, we test detectors with its testing or validation subset. We train each model 30K iterations for each epoch while the same amount of data are sampled from each training dataset to restrict the impact of difference of dataset sizes.

We evaluate detectors with the widely accepted criterion, mean Average Precision (mAP) and $MR^{-2}$ (i.e., log average miss rate over False Positive Per Image (FPPI) over range [$10^{-2}$, $10^0$]). Given occlusion is a key factor affecting the performance, we report different occlusion levels as shown in Table \ref{tab:occlusion}, following \cite{hasan2021generalizable}. Visible part is not labeled in WiderPedestrian, thus we report the overall $MR^{-2}$. More results are in Appendix.

\paragraph{Training details.}
We adopt ResNet-50 as the backbone of all methods. All of the methods are trained for 20 epochs and 30K iteration each epoch.
We follow the settings of hyperparameters presented in corresponding original papers, respectively. The initial learning rate is set to 0.02 and decayed by a factor of 10 after the 14th epoch and 18th epoch. We train all the methods on 8 GPUs and set the minibatch size to 16. The short edge of input images is resized to 1024 and the long edge is smaller than 1792. To reduce the impact of imbalance of data amount, we ensure that approximately the same number of images are sampled from each training datasets for each epoch.   

\subsection{Classic DG Evaluation in General Object Detection} 

We consider the classic DG evaluation method to evaluate detectors' generalization ability, where one dataset is selected for test and the others for training for each run. Note that for DG evaluation, knowledge of test distribution is completely inaccessible in the training phase, so that current detection methods designed for domain adaptation (DA) \cite{2018Cross,2021Generative,Guan2021UncertaintyAwareUD,Munir2021SSTNSD} are not applicable. Moreover, most current DG methods are designed for image classification and the adaptation of them in the object detection task is nontrivial \cite{Li2018LearningTG,Qiao2020LearningTL,Sun2016DeepCC}. 

Other than Faster R-CNN and RetinaNet, we select and reimplement the following model-agnostic DG methods as baselines. \textbf{Jigen\cite{Carlucci2019DomainGB}.} Jigen is a representative representation enhancement based method for DG without requirement of any extra annotations. We introduce an extra jiasaw classifier to Faster R-CNN and minimize the image-level jigsaw loss as suggested in the paper. \textbf{RSC.\cite{Huang2020SelfChallengingIC}} RSC is a dropout based DG method that iteratively discards the dominant features activated on the training data. \textbf{StableNet \cite{zhang2021deep}.} StableNet proposed to improve the generalization ability under distribution shifts via sample reweighting. We directly calculate RFF of image representations and adopt the image-wise reweighting in a Faster R-CNN model. 

These methods are easy to be assembled with object detection since they are not strongly coupled with the classification task and do not require domain labels.
The results are shown in Table \ref{tab:general}. The direct combinations of current DG methods and Faster R-CNN fail to achieve significant improvement, which may be caused by the two-stage optimization in Faster R-CNn and the small batch size (compared with the classification task, the batch size in object detection is considerably small). This further indicates that although the field of DGOB is of critical importance for real-world applications, it lacks competitive specially designed methods.

\subsection{Density Shift}
\begin{figure*}[t]
    \centering
    \includegraphics[width=0.84\linewidth]{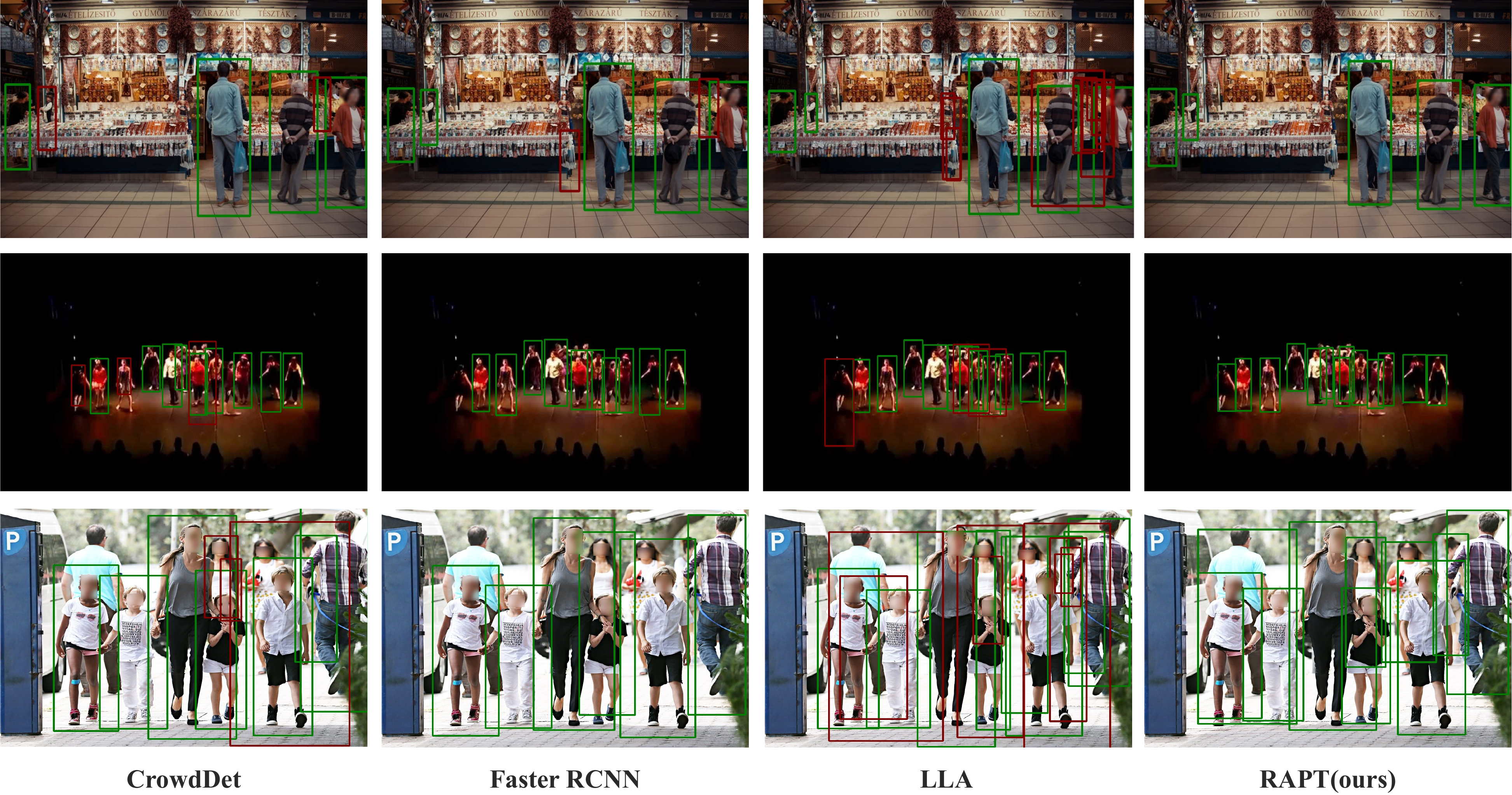}
    \caption{Visualization of bounding boxes detected by RAPT, CrowdDet, LLA and Faster RCNN. RAPT yields less wrong detected bounding boxes (marked with red anchors) compared with CrowdDet and LLA, and more accurate bounding boxes than Faster RCNN. }
    \label{fig:detec-case}
    \vspace{-3pt}
\end{figure*}

We investigate how the density shift between training and testing data affects current pedestrian detectors and the proposed method. Since pedestrian detection is widely used in many real-world applications where pedestrians can be excessively crowded (e.g., shopping malls, airports, and train stations) or quite sparse (e.g., streets and schoolyards), the generalization ability of detectors across people density is of great importance.  

Several previous works \cite{chu2020detection,hasan2021generalizable,2021LLA} have discussed the definition and impact of the instance density of pedestrian detection datasets under i.i.d. settings, yet none of them consider the distribution shift caused by the diversity of different datasets in instance density. 
We summarize the instance density of pedestrian detection datasets in Table \ref{tab:instance-density}.
We set the threshold of objects per image to 10 and cluster these datasets into dense ones, namely CrowdHuman, ECP, and sparse ones, namely WiderPedestrian, Caltech, and Citypersons. We first investigate the generalization ability of models trained on dense datasets and tested on sparse datasets. 

Results of detectors trained on dense scenes and tested on sparse scenes are shown in Table \ref{tab:dense1}. Surprisingly, we find that detectors specifically designed for pedestrian detection show worse performance compared with general object detectors. Even pedestrian detectors especially proposed for crowd scenarios such as CrowdDet\cite{chu2020detection} and IterDet\cite{DBLP:journals/corr/abs-2005-05708} fail to outperform general detectors such as Faster RCNN\cite{2017Faster}, indicating that though these methods show outstanding performance when tested in crowd scenarios, they fail to learn invariant features from data with different people densities. The proposed RAPT, on the contrary, achieves the best performance on almost all of the sub-settings. 

Results of detectors trained on sparse scenes and tested on dense ones are shown in Table \ref{tab:dense2}. Compared with Faster RCNN and RetinaNet\cite{2017Focal}, dense pedestrian detectors show superior performance. As a plug-in module, RAPT can be combined with any proposal-based detectors to improve their generalization ability under distribution shifts. We can easily introduce RAPT to CrowdDet via learning proposal weights for multiple proposals generated by CrowdDet and reweighting valid proposals to calculate final losses. We present results of RAPT with Faster RCNN and CrowdDet as the base model, respectively in Table \ref{tab:dense2}. RAPT further improves the performance of CrowdDet under dense testing scenarios.

\subsection{Context Shift}

Contexts of popular pedestrian detection benchmarks can be split into street scenarios and open-world scenarios. Datasets with street contexts, such as CityPersons, Caltech, and EuroCity Persons, can also be considered as autonomous driving datasets, while others, such as CrowdHuman and WiderPedestrian, are non-traffic related datasets. 

Investigating generalization ability from non-traffic-related data to autonomous driving data is of critical importance, given that pedestrian detection for autonomous driving is required to generate accurate predictions under any possible environments, which may exceed the coverage of training distribution. 
Moreover, data from open-world scenarios are more heterogeneous and easy to collect so that they can provide significant diversity of human features, such as various postures and occlusion situations. 
Thus the robustness of the model when adopted in unknown autonomous driving scenarios can be largely improved via non-traffic-related data.   

We first train our proposed method and state-of-the-art pedestrian detectors on non-traffic-related datasets before evaluating them on datasets with street scenarios context. Results are shown in Table \ref{tab:context1}. When testing on ECP, general detectors achieve better performance compared with pedestrian detectors, while ACSP\cite{DBLP:journals/corr/abs-2002-09053} shows advanced accuracy on \textsl{reasonable} and \textsl{small} sets of CityPersons. With raw Faster RCNN as the base model, RAPT shows the best performance on most of the settings, which indicates that the robustness of detectors for autonomous driving can be strengthened by RAPT with non-traffic data. Then we train detectors on traffic-related datasets and test them on the remaining ones and show the results in Table \ref{tab:context2}. RAPT consistently outperforms its state-of-the-art counterparts, showing clear improvements in generalization ability. 

\subsection{Random Shift}

We consider the random distribution shifts between training and testing data. Pedestrian detectors can encounter random or uncertain shifts other than pre-set ones such as \textsl{density shift} and \textsl{context shift}. We randomly select several datasets for training and others for evaluation to generate random distribution shifts. Here we present the results of training with ECP and CityPersons, testing with CrowdHuman, Caltech, and WiderPedestrian in Table \ref{tab:comp_random}. More experimental results under other splits of datasets are shown in Appendix. 

As shown in Table \ref{tab:comp_random}, RAPT outperforms other methods under random distribution shifts. Visualizations of bounding boxes detected by RAPT, CrowdDet, and Faster RCNN are shown in Figure \ref{fig:detec-case}. RAPT generates more accurate bounding boxes and fewer wrong bounding boxes, indicating a clear improvement of generalization ability under random distribution shifts.

\section{Conclusions}
In this paper, to investigate the impact of distribution shifts, we proposed three novel cross-dataset evaluation settings for pedestrian detectors. Then we proposed a novel method named RAPT, which can eliminate the statistical correlation between relevant and irrelevant features via learning proposal samples with RoI features, to improve the generalization of detection models under distribution shifts. Extensive experiments across several datasets and settings proved the effectiveness of our proposed method.


{\small

\bibliographystyle{ieee_fullname}
\bibliography{egbib}
}

\clearpage

\appendix
\onecolumn
\section{Appendix}

\subsection{Detectors Under Distribution Shifts}

\begin{table*}[ht]
    \centering
    \caption{$MR^{-2}$ of detectors trained and tested with the same dataset. All detectors are reimplemented with ResNet-50 \cite{he2016deep} pretrained on Imagenet \cite{deng2009imagenet} as the backbone. The title of each column indicates the tested subset. The best results of all methods are highlighted with the bold font.}
    \resizebox{0.85\textwidth}{!}{
    \begin{tabular}{c|cccc|cccc|c}
        \toprule
         \multirow{2}{*}{Model} &  \multicolumn{4}{c|}{ECP} & \multicolumn{4}{c|}{CrowdHuman} & \multicolumn{1}{c}{WiderPedestrian} \\
         \cmidrule{2-10} 
          & Reasonable & Small & Heavy & All & Reasonable & Small & Heavy & All & All \\
        \midrule
        CrowdDet & \textbf{0.0941} & \textbf{0.1442} & \textbf{0.3899} & \textbf{0.3322} & \textbf{0.1771} & \textbf{0.1835} & \textbf{0.4056} & \textbf{0.3884} & \textbf{0.5329} \\
        LLA  & 0.1387 & 0.2093 & 0.4701 & 0.4033 & 0.1944 & 0.2092	& 0.4287 & 0.4096 & 0.5774  \\
        RetinaNet  & 0.1674 & 0.2584 & 0.4817 & 0.4692 & 0.2976 & 0.2830 & 0.4907	& 0.5418 & 0.5556\\
        Faster RCNN  & 0.1273 & 0.2017 & 0.4616 & 0.4071 & 0.1912 & 0.2008 & 0.4226 & 0.4025 & 0.5617  \\
        RAPT (ours) + FRCNN  & 0.1235 & 0.2253 &
        0.4675 & 0.4109 & 0.1920 & 0.2115 & 0.4157 & 0.4079 & 0.5596\\
        \bottomrule
    \end{tabular}}
    \label{tab:single1}
\end{table*}

As discussed in Section 4, despite the striking performance current pedestrian detectors achieved when trained and tested with a single dataset, they suffer a significant drop under distribution shifts between training and testing data. Here we present the results of detectors trained and tested with a single dataset in Table \ref{tab:single1}. Compared with results in Table 1, Table 4 and Table 5 in the main paper, detectors trained and tested with a single dataset show significant higher performance on ECP, CrowdHuman and WiderPedestrian, which indicating clear performance drop caused by distribution shifts between training and testing datasets. The results meet observations in previous works \cite{hasan2021generalizable, 2021Robust} and show that distribution shifts is a crucial problem for the real-world applications of pedestrian detectors.
Furthermore, RAPT shows no superior performance compared to current detectors, yet shows considerable improvement under distribution shifts.

\subsection{More Experimental Results}
\begin{table*}[ht]
    \centering
    \caption{mean Average Precision (mAP) of detectors trained on CrowdHuman and ECP, and tested on CityPersons, Caltech and WiderPedestrian. All detectors are reimplemented with ResNet-50 \cite{he2016deep} pretrained on Imagenet \cite{deng2009imagenet} as the backbone. 
     The title of each column indicates the tested subset. The best results of all methods are highlighted with the bold font.}
    \resizebox{\textwidth}{!}{
    \begin{tabular}{c|c|cccc|cccc|cccc}
        \toprule
         Method & Training Data & \multicolumn{4}{c|}{CityPersons} & \multicolumn{4}{c|}{Caltech}  & \multicolumn{4}{c}{WiderPedestrian} \\
         \cmidrule{3-14} 
         & & Small & Medium & Large & All & Small & Medium & Large & All & Small & Medium & Large & All  \\
        \midrule
    
        CrowdDet \cite{chu2020detection} & CrowdHuman+ECP & 0.101 & 0.406 & 0.519 & 0.368 & 0.080 & 0.283 & 0.450 & 0.169 & 0.098 & 0.333 & 0.495 & 0.340\\
        IterDet  \cite{DBLP:journals/corr/abs-2005-05708} & CrowdHuman+ECP & 0.068 & 0.372 & 0.463 & 0.328 & 0.041 & 0.213 & 0.407 & 0.119 & 0.061 & 0.242 & 0.339 & 0.231 \\
        ACSP \cite{DBLP:journals/corr/abs-2002-09053} & CrowdHuman+ECP & 0.050 & 0.390 & 0.590 & 0.367 & 0.049 & 0.229 & 0.415 & 0.126 & 0.067 & 0.251 & 0.355 & 0.249 \\
        LLA \cite{2021LLA} & CrowdHuman+ECP & 0.132 & 0.397 & 0.486 & 0.362 & 0.089 & 0.286 & \textbf{0.454} & 0.174 & 0.105 & 0.326 & 0.476 & 0.330 \\
        RetinaNet \cite{2017Focal} & CrowdHuman+ECP & 0.072 & 0.387 & 0.484 & 0.342 & 0.088 & 0.297 & 0.434 & 0.177 & 0.105 & 0.334 & 0.492 & 0.336\\
        Cascade RCNN \cite{0Cascade} & CrowdHuman+ECP & 0.070 & 0.378 & 0.472 & 0.339 & 0.042 & 0.217 & 0.412 & 0.123 & 0.062 & 0.247 & 0.348 & 0.238 \\
        \midrule
        Faster RCNN \cite{2017Faster} & CrowdHuman+ECP & \textbf{0.160} & 0.425 & 0.503 & 0.389 & 0.101 & 0.301 & 0.441 & 0.187 & \textbf{0.127} & 0.348 & \textbf{0.513} & 0.358\\
        RAPT (ours) + FRCNN & CrowdHuman+ECP  & 0.156 & \textbf{0.433} & \textbf{0.515} & \textbf{0.403} & \textbf{0.112}  & \textbf{0.325} & 0.437 & \textbf{0.193} & 0.112 & \textbf{0.353} & 0.499 & \textbf{0.368} \\
        \bottomrule
    \end{tabular}}
    \label{tab:map1}
\end{table*}

\begin{table*}[ht]
    \centering
    \caption{Mean Average Precision (mAP) of detectors trained on open-world datasets and tested on autonomous driving ones. For details about the number of runs, meaning of column titles and fonts, see Table \ref{tab:single1}.}
    \resizebox{\textwidth}{!}{
    \begin{tabular}{c|c|cccc|cccc|cccc}
        \toprule
         \multirow{2}{*}{Model} & \multirow{2}{*}{Training} & \multicolumn{4}{c|}{ECP} & \multicolumn{4}{c|}{Caltech}  & \multicolumn{4}{c}{CityPersons} \\
         \cmidrule{3-14} 
         & & Small & Medium & Large & All & Small & Medium & Large & All & Small & Medium & Large & All  \\
        \midrule
        CrowdDet & C+W & 0.099 & 0.349 & \textbf{0.555} & 0.322 & 0.073 & 0.276 & 0.449 & 0.165 & 0.067 & 0.320 & 0.487 & 0.306 \\
        IterDet & C+W & 0.062 & 0.293 & 0.387 & 0.246 & 0.057 & 0.239 & 0.413 & 0.138 & 0.052 & 0.312 & 0.451 & 0.289 \\
        LLA & C+W & \textbf{0.124} & 0.352 & 0.538 & 0.328 & 0.072 & 0.255 & 0.429 & 0.156 & \textbf{0.087} & 0.321 & 0.449 & 0.301 \\
        RetinaNet & C+W & 0.096 & 0.354 & 0.548 & 0.321 & 0.073 & 0.270 & 0.421 & 0.161 & 0.057 & 0.324 & 0.463 & 0.298 \\
        Cascade RCNN & C+W & 0.071 & 0.302 & 0.393 & 0.255 & 0.061 & 0.246 & 0.422 & 0.144 & 0.056 & 0.318 & 0.458 & 0.294 \\
        \midrule
        Faster RCNN & C+W & 0.104 & 0.357 & 0.549 & 0.326 & \textbf{0.078} & 0.271 & 0.438 & 0.162 & 0.067 & 0.330 & 0.477 & 0.308 \\
        RAPT (ours) + FRCNN & C+W  & 0.115 & \textbf{0.375} & 0.539 & \textbf{0.340} & 0.070 & \textbf{0.279} & \textbf{0.446} & \textbf{0.172} & 0.076 & \textbf{0.351} & \textbf{0.489} & \textbf{0.320} \\
        \bottomrule
    \end{tabular}}
    \label{tab:map2}
\end{table*}

We present mean Average Precision (mAP) of detectors trained on CrowdHuman and ECP, and tested on CityPersons, Caltech and WiderPedestrian in Table \ref{tab:map1}, which shares the same experimental setting with Table 1 in the main paper. We present mean Average Precision (mAP) of detectors trained on CrowdHuman and WiderPedestrian, and tested on ECP, Caltech and CityPersons in Table \ref{tab:map2}, which shares the same experimental setting with Table 5 in the main paper.
The proposed RAPT consistently shows superior mAP on all datasets.

\begin{table*}[ht]
    \centering
    \caption{$MR^{-2}$ of detectors evaluated under random distribution shifts. For details about the number of runs, meaning of column titles and fonts, see Table \ref{tab:single1}.}
    \resizebox{0.8\textwidth}{!}{
    \begin{tabular}{c|c|cccc|c|c}
        \toprule
        \multirow{2}{*}{Model} & \multirow{2}{*}{Training Data} & \multicolumn{4}{c|}{CrowdHuman} &  \multicolumn{1}{c|}{WiderPedestrian} & \multicolumn{1}{c}{COCO} \\
        \cmidrule{3-8} 
        & & Reasonable & Small & Heavy & All & All & All  \\
        \midrule
        CrowdDet & ECP+Caltech & 0.6806 & 0.6118 & 0.9188 & 0.8367 & 0.8153 & 0.8051 \\
        LLA & ECP+Caltech & \textbf{0.6548} & \textbf{0.6073} & 0.9221 & 0.8350  & 0.8151 & \textbf{0.7715} \\
        RetinaNet & ECP+Caltech & 0.6968 & 0.6385 & 0.9280 & 0.8580 & 0.8132  & 0.7744 \\
        \midrule
        Faster RCNN & ECP+Caltech & 0.6740 & 0.6286 & 0.9242 & 0.8268 & 0.8017 & 0.7927 \\
        RAPT (ours)+FRCNN & ECP+Caltech  & 0.6651 & 0.6108 & \textbf{0.9003} & \textbf{0.8173} &  \textbf{0.7981}  & 0.7825  \\
        \bottomrule
    \end{tabular}}
    \label{tab:comp_random-appendix}
\end{table*}

\subsection{More Bounding Boxes Visualization.}
\begin{figure*}[h]
    \centering
    \includegraphics[width=0.85\linewidth]{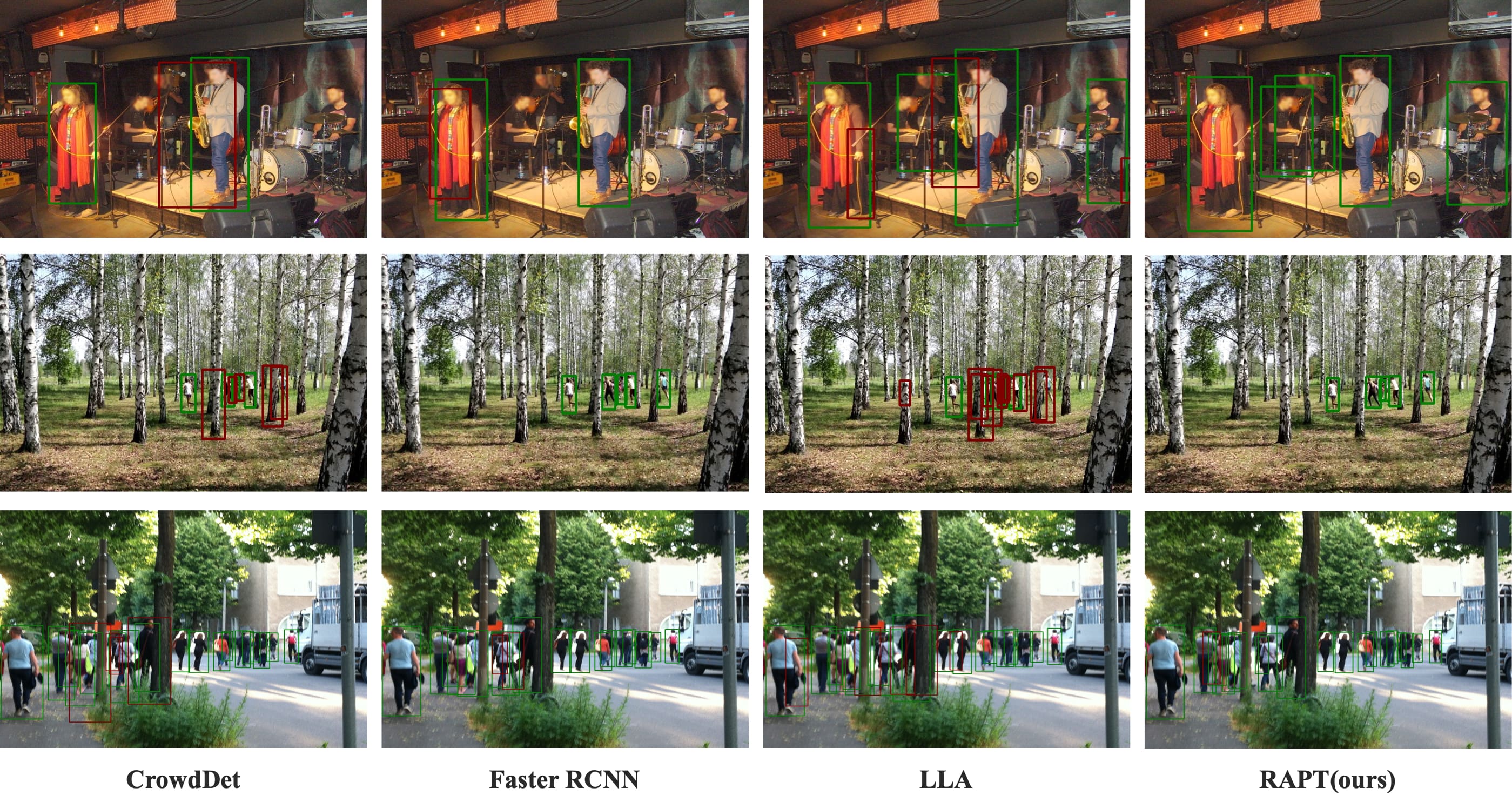}
    \caption{Visualization of bounding boxes detected by RAPT, CrowdDet, LLA and Faster RCNN. RAPT yields less wrong detected bounding boxes (marked with red anchors) compared with CrowdDet and LLA, and more accurate bounding boxes than Faster RCNN. }
    \label{fig:vis}
    \vspace{-3pt}
\end{figure*}

\begin{figure*}[t]
    \centering
    \includegraphics[width=0.85\linewidth]{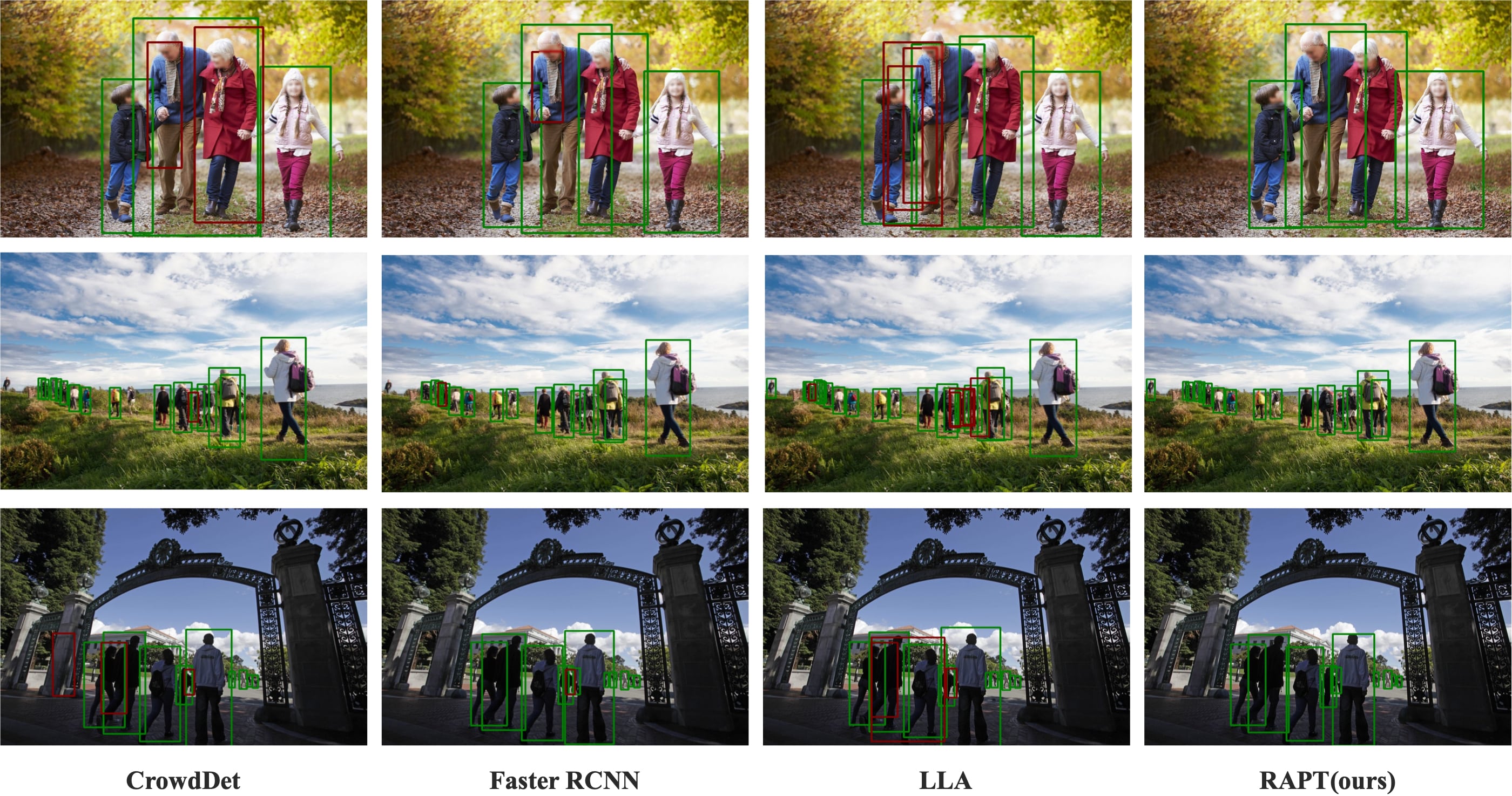}
    \includegraphics[width=0.85\linewidth]{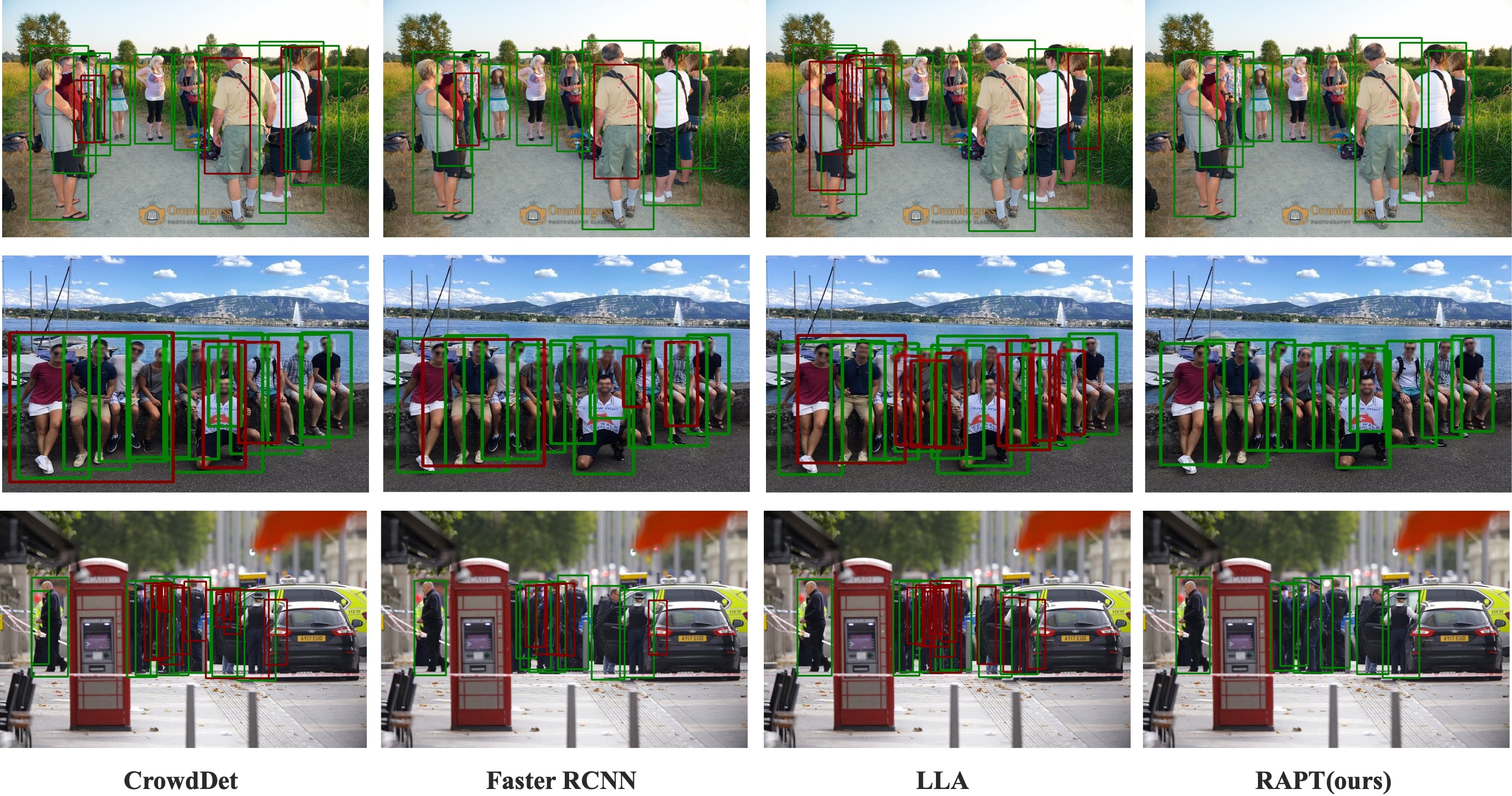}
    \caption{Visualization of bounding boxes detected by RAPT, CrowdDet, LLA and Faster RCNN. RAPT yields less wrong detected bounding boxes (marked with red anchors) compared with CrowdDet and LLA, and more accurate bounding boxes than Faster RCNN. }
    \label{fig:vis2}
    \vspace{-3pt}
\end{figure*}

We show more bounding boxes visualization in Figure \ref{fig:vis} and Figure \ref{fig:vis2}. RAPT generates more accurate bounding boxes and fewer wrong bounding boxes, indicating a clear improvement of generalization ability under distribution shifts.

\end{document}